\PassOptionsToPackage{warn}{textcomp}
\documentclass[runningheads]{llncs}
\pdfoutput=1
\usepackage[section]{placeins}
\usepackage{times}
\usepackage{latexsym}

\usepackage{subcaption}
\usepackage{graphicx}
\usepackage{xcolor}
\usepackage{url}
\usepackage{soul}
\usepackage[T2A,T1]{fontenc}
\usepackage[utf8]{inputenc}
\usepackage[russian,english]{babel}
\usepackage{csquotes}
\usepackage{amsmath}
\usepackage{amssymb}
\usepackage[warn]{textcomp}
\usepackage{tabularx}
\usepackage{romannum}
\usepackage{fixltx2e}
\mathchardef\mhyphen="2D

\AtBeginDocument{\pagenumbering{arabic}}
\begin{document}

\title{Studying Attention Models  in \\ Sentiment Attitude Extraction Task\thanks{The reported study was partially supported by RFBR, research project \textnumero~20-07-01059}}
\titlerunning{Studying Attention Models in Sentiment Attitude Extraction Task}

\author{Nicolay Rusnachenko\inst{1} \and
Natalia Loukachevitch\inst{1,2}}

\authorrunning{N. Rusnachenko and N. Loukachevitch}

\institute{
Bauman Moscow State Technical University, Moscow, Russia \\
\email{kolyarus@yandex.ru}
\and
Lomonosov Moscow State University, Moscow, Russia \\
\email{louk\_nat@mail.ru}
}

\maketitle

\newcommand{\rusentrel}{RuSentRel}
\newcommand{\devDataset}{RuAttitudes}
\newcommand{\rusentiframes}{RuSentiFrames}
\newcommand{\rusentilex}{RuSentiLex}

\newcommand{\test}{\textsc{test}}
\newcommand{\train}{\textsc{train}}
\newcommand{\testSc}{\textsubscript{\test{}}}
\newcommand{\trainSc}{\textsubscript{\train{}}}


\newcommand{\improveRange}{$1.5$-$5.9$}
\newcommand{\improvePZhou}{$9.8$}
\newcommand{\epochsToTest}{10}
\newcommand{\epochsCount}{150}
\newcommand{\pcnnIncrease}{1.4}
\newcommand{\ianIncrease}{1.2}
\newcommand{\bilstmIncrease}{5.9}

\newcommand{\maskE}{$E$}
\newcommand{\maskEObj}{$\underline{E}_{obj}$}
\newcommand{\maskESubj}{$\underline{E}_{subj}$}

\newcommand{\argZero}{\texttt{Arg0}}
\newcommand{\argOne}{\texttt{Arg1}}
\newcommand{\polarity}{\texttt{A0}$\to$\texttt{A1}}

\newcommand{\featuresCount}{k}
\newcommand{\embSize}{m}
\newcommand{\ctxSize}{n}
\newcommand{\labelsCount}{c}

\newcommand{\word}[1]{t_{#1}}
\newcommand{\embWord}[1]{x_{#1}}
\newcommand{\embFeature}[1]{f_{#1}}
\newcommand{\hidden}[1]{h_{#1}}
\newcommand{\entity}[1]{e_{#1}}

\newcommand{\featureSet}{F}
\newcommand{\embSet}{X}

\newcommand{\sentEmbSize}{z}

\newcommand{\entityGroup}{\textsc{entities}}
\newcommand{\wordsGroup}{\textsc{words}}

\newcommand{\han}{\textsc{HAN}}
\newcommand{\lstm}{\textsc{LSTM}}

\newcommand{\blstm}{\textsc{B}\lstm{}}
\newcommand{\bilstm}{\textsc{B}i\lstm{}}
\newcommand{\bilstmPZhou}{\textsc{Att}-\blstm{}}
\newcommand{\bilstmZYang}{\textsc{Att}-\blstm{}$^{z\mhyphen yang}$}

\newcommand{\featureEnds}{${{att\mhyphen ends}}$}
\newcommand{\featureFrames}{${{att\mhyphen frames}}$}
\newcommand{\featureEf}{${{att\mhyphen ef}}$}

\newcommand{\attCnn}{\textsc{AttCNN}}

\newcommand{\ian}{\textsc{IAN}}
\newcommand{\ianEnds}{\ian{}$_{{ends}}$}
\newcommand{\ianFrames}{\ian{}$_{{frames}}$}
\newcommand{\ianSef}{\ian{}$_{{sef}}$}
\newcommand{\ianSe}{\ian{}$_{{se}}$}
\newcommand{\ianEf}{\ian{}$_{{ef}}$}

\newcommand{\pcnn}{\textsc{PCNN}}
\newcommand{\pcnnEnds}{\pcnn{}$_{{att\mhyphen ends}}$}
\newcommand{\pcnnFrames}{\pcnn{}$_{{att\mhyphen frames}}$}
\newcommand{\pcnnEf}{\pcnn{}$_{{att\mhyphen ef}}$}

\newcommand{\cnn}{\textsc{CNN}}
\newcommand{\cnnEnds}{\cnn{}$_{{att\mhyphen ends}}$}
\newcommand{\cnnFrames}{\cnn{}$_{{att\mhyphen frames}}$}
\newcommand{\cnnEf}{\cnn{}$_{{att\mhyphen ef}}$}

\newcommand{\devDatasetShort}{RA}
\newcommand{\rusentrelShort}{RSR}

\newcommand{\expA}{\devDatasetShort{}}
\newcommand{\expB}{\rusentrelShort{}}
\newcommand{\expC}{\expB{}+\devDatasetShort{}}

\newcommand{\tokenGroup}{\textsc{tokens}}
\newcommand{\prepGroup}{\textsc{prep}}
\newcommand{\framesGroup}{\textsc{frames}}
\newcommand{\sentGroup}{\textsc{sentiment}}

\newcommand{\precision}{$P$}
\newcommand{\recall}{$R$}
\newcommand{\fmeasure}{$F1$}

\newcommand{\stemmer}{\url{https://tech.yandex.ru/mystem/}}

\newcommand{\rusentrelLink}{\url{https://github.com/nicolay-r/RuSentRel/tree/v1.1}}
\newcommand{\ruSentiFramesLink}{\url{https://github.com/nicolay-r/RuSentiFrames/tree/v1.0}}
\newcommand{\devDatasetLink}{\url{https://github.com/nicolay-r/RuAttitudes/tree/v1.0}}
\newcommand{\experimentsLink}{\url{https://github.com/nicolay-r/attitudes-extraction-ds}}

\newcommand{\bagsCount}{l}
\newcommand{\sentencesCount}{t}
\newcommand{\classesCount}{c}
\newcommand{\sentenceEmbedding}{s}
\newcommand{\sentenceEmbeddingSet}{E_{s}}

\begin{abstract}
In the sentiment attitude extraction task, the aim is to identify 
<<attitudes>> -- sentiment relations between entities mentioned 
in text. 
In this paper, we provide a study on attention-based context encoders in the
sentiment attitude extraction task.
For this task, we adapt attentive context encoders of two types:
(\romannum{1}) feature-based;
(\romannum{2}) self-based.
Our experiments\footnote{\url{https://github.com/nicolay-r/attitude-extraction-with-attention}}
with a corpus of Russian analytical texts 
\rusentrel{} illustrate  that the models trained with attentive 
encoders outperform ones that were trained without them
and achieve \improveRange{}\% increase by \fmeasure{}. We also provide the analysis of attention weight distributions in dependence on the term type.

\end{abstract}
\section{Introduction}


    Classifying relations between entities mentioned in texts remains one of the popular tasks in natural language processing (NLP).
    The sentiment attitude extraction task aims to 
    seek for positive/negative relations between  objects expressed as named entities  in texts~\cite{rusnachenko2018neural}.
    Let us consider  the following sentence as an example (named entities are underlined):

\begin{center}
    ``Meanwhile \underline{Moscow} has repeatedly emphasized that its activity in the
    \underline{Baltic Sea} is a response precisely to actions of
    \underline{NATO}
    and the escalation of the hostile approach to \underline{Russia} near its eastern borders''
\end{center}

    In the example above,  named entities <<Russia>> and <<NATO>> have the  negative attitude towards each other
    with additional indication of other named entities.
    The complexity of the sentence structure is one of the greatest difficulties one encounters 
    when dealing with the relation extraction  task.
    Texts usually contain a lot of named entity mentions;
    a single opinion might comprise  several sentences.

    This paper is devoted to study  of models for targeted sentiment analysis with attention.
    The intuition exploited in the models with attentive encoders  is that not all terms in the context are relevant for attitude indication.
    The interactions of  words, not just their isolated presence, may reveal the specificity of 
    contexts with attitudes of different polarities.
    The primary contribution of this work is an application of attentive encoders based on
    (\romannum{1}) sentiment  frames and attitude participants (features);
    (\romannum{2}) context itself.
    We conduct  the experiments on the  \rusentrel{}~\cite{loukachevitch2016creating} collection.
    The results demonstrate that attentive  models  with \cnn{}-based and over \lstm{}-based encoders
    result in \improveRange{}\% by \fmeasure{} over models without attentive encoders.


\section{Related Work}

In previous works, various neural network approaches for targeted sentiment analysis were proposed.
In~\cite{rusnachenko2018neural} the authors utilize 
convolutional neural networks (\cnn{}).
Considering  relation extraction as a three-scale classification task of contexts with attitudes in it,
the authors subdivide each context into \textit{outer} and \textit{inner} 
(relative to attitude participants) to apply Piecewise-\cnn{} (\pcnn{})~\cite{zeng2015distant}.
The latter architecture utilizes a specific idea of \textit{max-pooling} operation.
Initially, this is an operation, which  extracts 
the maximal values within each convolution.
However, for relation classification, it reduces information extremely rapid and blurs 
significant aspects of 
context parts. In case of PCNN,   separate max-pooling operations are applied to outer and inner contexts. 
In the experiments, the authors revealed a fast training process and a slight improvement in the \pcnn{} results  in comparison to \cnn{}.

In~\cite{shen-huang-2016-attention}, the authors proposed an attention-based CNN model for semantic relation classification~\cite{hendrickx2009semeval}.
The authors utilized the attention mechanism to select the most relevant context words with respect to participants of a  semantic relation.
The architecture of the attention model is a multilayer perceptron (MLP), which  calculates the weight of a word in context with respect to the entity.
The resulting \textsc{AttCNN} model outperformed several \textsc{CNN} and \textsc{LSTM} based approaches with $2.6\mhyphen{}3.8\%$ by F1-measure.

In~\cite{ma2017interactive},
the authors experimented with attentive models in aspect-based sentiment analysis.
The  models were aimed to identify sentiment polarity of specific \textit{targets} in context, which are characteristics or parts of an entity.
Both targets and the context were treated as \textit{sequences}.
The authors proposed an interactive attention network (\ian{}), 
which establishes element relevance of one sequence with the other in two directions:
targets to context,
context to targets.
The effectiveness of \textsc{IAN} was demonstrated on the SemEval-2014 dataset~\cite{wagner2014dcu}
and several biomedical datasets~\cite{alimova2018interactive}.

In \cite{zhou2016attention,yang2016hierarchical}, the authors experimented with self-based attention models, in which \textit{targets} became adapted automatically during the training process. 
Comparing with \ian{},
the presence of targets might be unclear in terms of algorithms. 
The authors considered the attention as context word quantification with respect to abstract targets.
In \cite{yang2016hierarchical}, the authors brought a similar idea also onto the sentence level. 
The obtained hierarchical model was called as \han{}.


\section{Data and Lexicons}
\label{sec:data}

We consider sentiment analysis of Russian analytical articles collected in the RuSentRel corpus~\cite{loukachevitch2018extracting}.
The corpus comprises texts in the international politics domain and contains a lot of opinions.
The articles are labeled with  annotations of two types:
(\romannum{1}) the author's opinion on the subject matter of the article;
(\romannum{2}) the attitudes between the participants of the described situations.
The annotation of the latter type includes 2000 relations across 73 large analytical texts.
Annotated sentiments can be only \textit{positive} or \textit{negative}.
Additionally,  each text is provided  with annotation of mentioned named entities. Synonyms and variants of named entities are also given, which allows not to deal with the coreference of named entities.

In our study, we also use two Russian sentiment resources: the \rusentilex{} lexicon~\cite{loukachevitch2016creating},
which contains words and expressions of the Russian language with sentiment labels and the \rusentiframes{} lexicon~\cite{rusnachenko2019distant},
which provides several types of sentiment attitudes for situations associated with specific Russian predicates.


The \rusentiframes{}\footnote{\ruSentiFramesLink{}} lexicon describes sentiments and connotations conveyed with a predicate in a verbal or nominal form  \cite{rusnachenko2019distant}, such as \begin{otherlanguage*}{russian}"осудить, улучшить, преувеличить" \end{otherlanguage*} (to condemn, to improve, to exaggerate), etc.
The structure of the frames in RuSentFrames comprises:
(\romannum{1}) the set of predicate-specific roles;
(\romannum{2}) frames dimensions such as the attitude of the author towards participants of the situation, attitudes between the participants, effects for participants. Currently, RuSentiFrames contains frames for more than 6 thousand  words and expressions.

In RuSentiFrames, individual  semantic roles are numbered, beginning with zero. For a particular predicate entry, \argZero{} is generally the argument exhibiting features of a Prototypical Agent, while \argOne{} is a Prototypical Patient or Theme \cite{dowty1991thematic}.
In the main part of the frame, the most applicable for the current study is 
the polarity of \argZero{} with a respect to \argOne{} (\polarity{}).
For example, in case of Russian verb
\begin{otherlanguage*}{russian} "одобрить" \end{otherlanguage*} (to approve)
the sentiment polarity \polarity{} is positive.

\section{Model}
\label{sec:model}

In this paper, the task of sentiment attitude extraction is treated as follows:
given  a pair of  named entities, we predict a sentiment label of
a pair, which could be positive, negative, or \textit{neutral}.
As the \rusentrel{} corpus provides opinions with positive or negative sentiment labels only~(Section~\ref{sec:data}),
we automatically added neutral sentiments for all pairs not
mentioned in the annotation and co-occurred in the same sentences of the collection texts. We consider a \textit{context} as a text fragment that is limited by a single sentence and includes a pair of named
entities.


\begin{figure}[t]
    \includegraphics[width=0.40\linewidth]{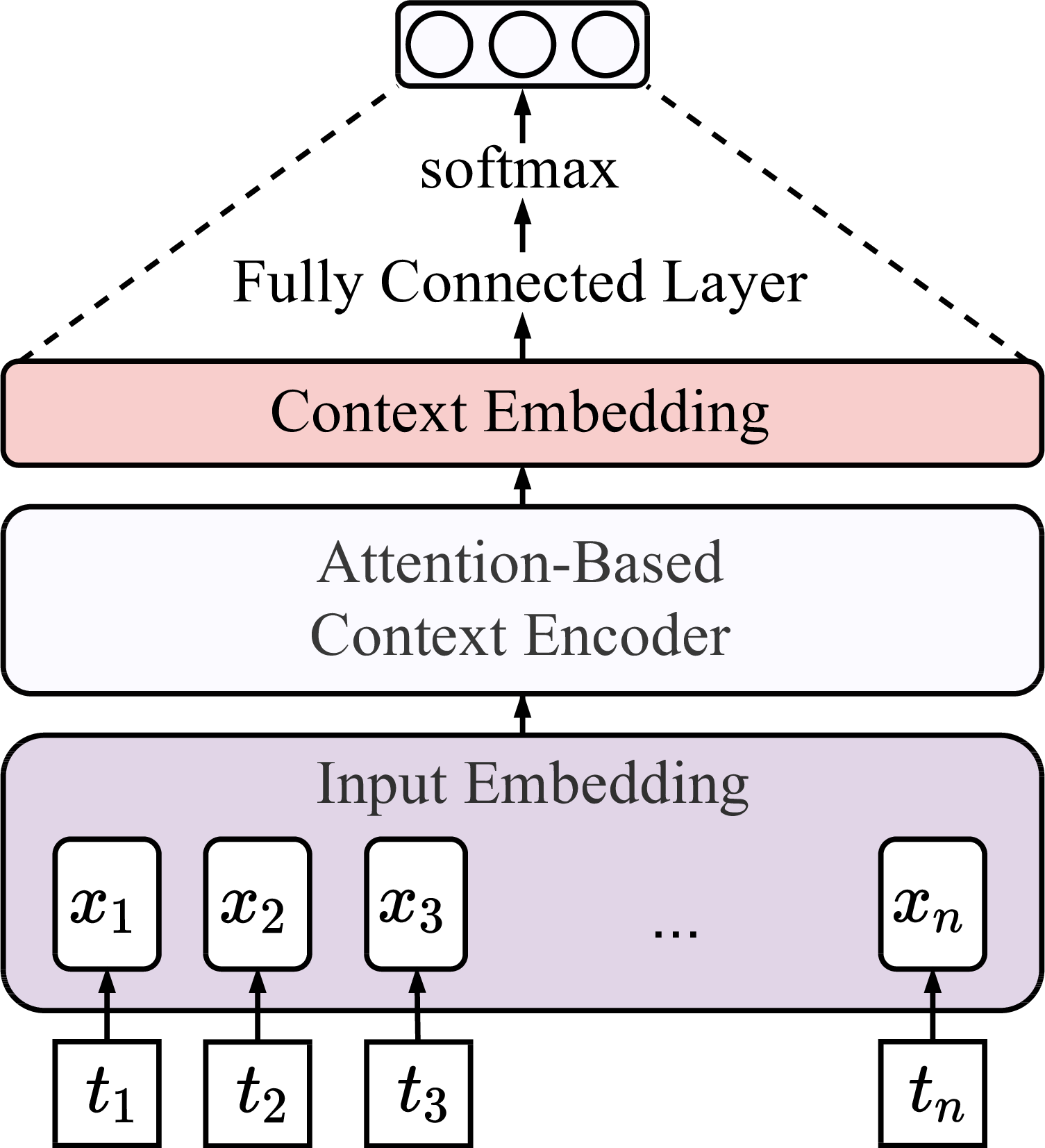}~\hspace{0.2cm}
    \includegraphics[width=0.571\linewidth]{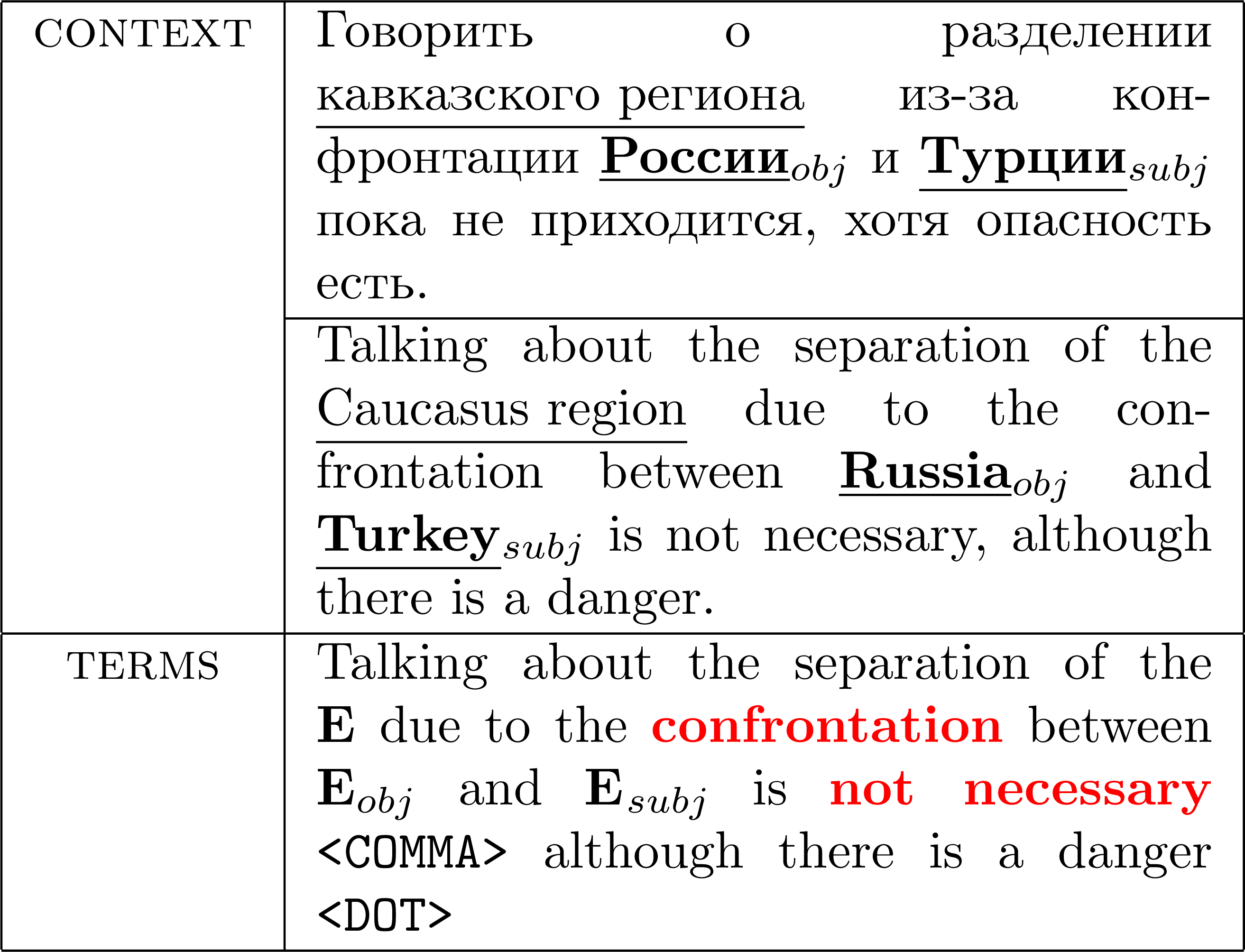}
    \caption{
     (\textit{left}) General, context-based 3-scale (positive, negative, neutral) classification model,
     with details on <<Attention-Based Context Encoder>> block in Section~\ref{sec:feature-based}~and~\ref{sec:self-based};
     (\textit{right}) An example of a context processing into a sequence of terms;
     attitude participants (<<Russia>>, <<Turkey>>) and other mentioned entities become masked;
     frames are bolded and optionally colored corresponding to the sentiment value of \polarity{} polarity.
    }
    \label{fig:model}
\end{figure}

The general architecture is presented in Figure~\ref{fig:model} (left), where
the sentiment could be extracted from the context.
To present a context, we treat the original text as a sequence of terms
$[\word{1}, \ldots, \word{\ctxSize{}}]$ limited by $\ctxSize{}$.
Each term belongs to one of the following classes:
\entityGroup{},
\framesGroup{}, \tokenGroup{}, and
\wordsGroup{} (if none of the prior has not been matched).
We use masked representation for attitude participants (\maskEObj{}, \maskESubj{})
and mentioned named entities (\maskE{}) to prevent models from capturing related information.

To represent \framesGroup{}, we combine 
a frame entry with the corresponding \polarity{} sentiment polarity value
(and \textit{neutral} if the latter is absent).
We also invert sentiment polarity when an entry has
\begin{otherlanguage*}{russian}"не"\end{otherlanguage*}
(not) preposition.
For example, in Figure~\ref{fig:model} (right) all entries
are encoded with the negative polarity \polarity{}:
\begin{otherlanguage*}{russian}"конфронтация"\end{otherlanguage*} (confrontation)
has a negative polarity, and
\begin{otherlanguage*}{russian}"не приходится"\end{otherlanguage*} (not necessary)
has a positive polarity of entry "necessary" which is inverted due to the "not" preposition.

The~\tokenGroup{} group includes: punctuation marks, numbers, url-links. 
Each term of \wordsGroup{} is considered in a lemmatized\footnote{\stemmer{}} form.
Figure~\ref{fig:model} (right) provides a context example with the
corresponding representation (<<\textsc{terms}>> block).

To represent the context in a model, each term is embedded with a 
vector of fixed dimension.
The sequence of embedded vectors $\embSet{} = [\embWord{1}, \ldots, \embWord{\ctxSize{}}]$ is
denoted as \textit{input embedding} ($\embWord{i} \in \mathbb{R}^\embSize{}, i \in \overline{1..\ctxSize{}}$).
Sections~\ref{sec:feature-based}~and~\ref{sec:self-based} provide an encoder implementation in details.
In particular, each encoder relies on input embedding and generates
output \textit{embedded context} vector $\sentenceEmbedding{}$. 

In order to determine a sentiment class by the embedded context $\sentenceEmbedding{}$, 
we apply:
(\romannum{1})~the hyperbolic tangent activation function towards $\sentenceEmbedding{}$ and
(\romannum{2})~transformation through the \textit{fully connected layer}:

\newcommand{\WHidden}{W_r}
\newcommand{\BHidden}{b_r}

\begin{equation}
    r = \WHidden{} \cdot \tanh(s) + \BHidden{}
    \hspace{1cm} 
    \WHidden{} \in \mathbb{R}^{\sentEmbSize{} \times \labelsCount{}},
    \hspace{0.2cm} 
    \BHidden{} \in \mathbb{R}^\labelsCount{},
    \hspace{0.2cm} \labelsCount{} = 3
    \label{eq:fc}
\end{equation}

\newcommand{\smProb}{\rho}

In Formula~\ref{eq:fc}, $\WHidden{}, \BHidden{}$ corresponds to hidden states;
$\sentEmbSize{}$ correspond to the size of vector $\sentenceEmbedding{}$,
and $\labelsCount{}$ is a number of classes.
Finally, to obtain an output vector of probabilities $o=\{\smProb{}_i\}_{i=1}^c$,
we use $softmax$ operation:
\begin{equation}
    \smProb{}_i = softmax(r_i) = \frac{\exp(r_i)}{\sum_{j=1}^{c}{\exp(r_j)}}
    \label{eq:softmax}
\end{equation}
\subsection{Feature Attentive Context Encoders}
\label{sec:feature-based}


In this section, we consider \textit{features} as a
significant for attitude identification context terms, towards which we would like
to quantify the relevance of each term in the context.
For a particular context, we select embedded values of the
(\romannum{1})~attitude participants (\maskEObj{}, \maskESubj{}) and
(\romannum{2})~terms  of the \framesGroup{} group
and create a set of features $\featureSet{} = [\embFeature{1}, \ldots, \embFeature{\featuresCount{}}]$
limited by~$\featuresCount{}$.

\subsubsection*{MLP-Attention.}

\newcommand{\featEmb}{\hat{s}}
\newcommand{\mlpHidden}{\textbf{h}_\textsc{mlp}}
\newcommand{\convCount}{\textbf{c}}


Figure~\ref{fig:mlp-based-model} illustrates a feature-attentive encoder with the quantification approach called Multi-Layer Perceptron~\cite{huang2016attention}.
In formulas~\ref{eq:mlp_concat}--\ref{eq:mlp_emb}, we
describe the quantification process  of a context embedding $\embSet{}$ with respect to a particular
feature $\embFeature{}~\in~\featureSet{}$.
Given an $i$'th embedded term $\embWord{i}$,
we concatenate its representation with $\embFeature{}$:

\begin{figure}[!t]
    \begin{subfigure}{.59\textwidth}
        \centering
        \includegraphics[width=0.95\textwidth]{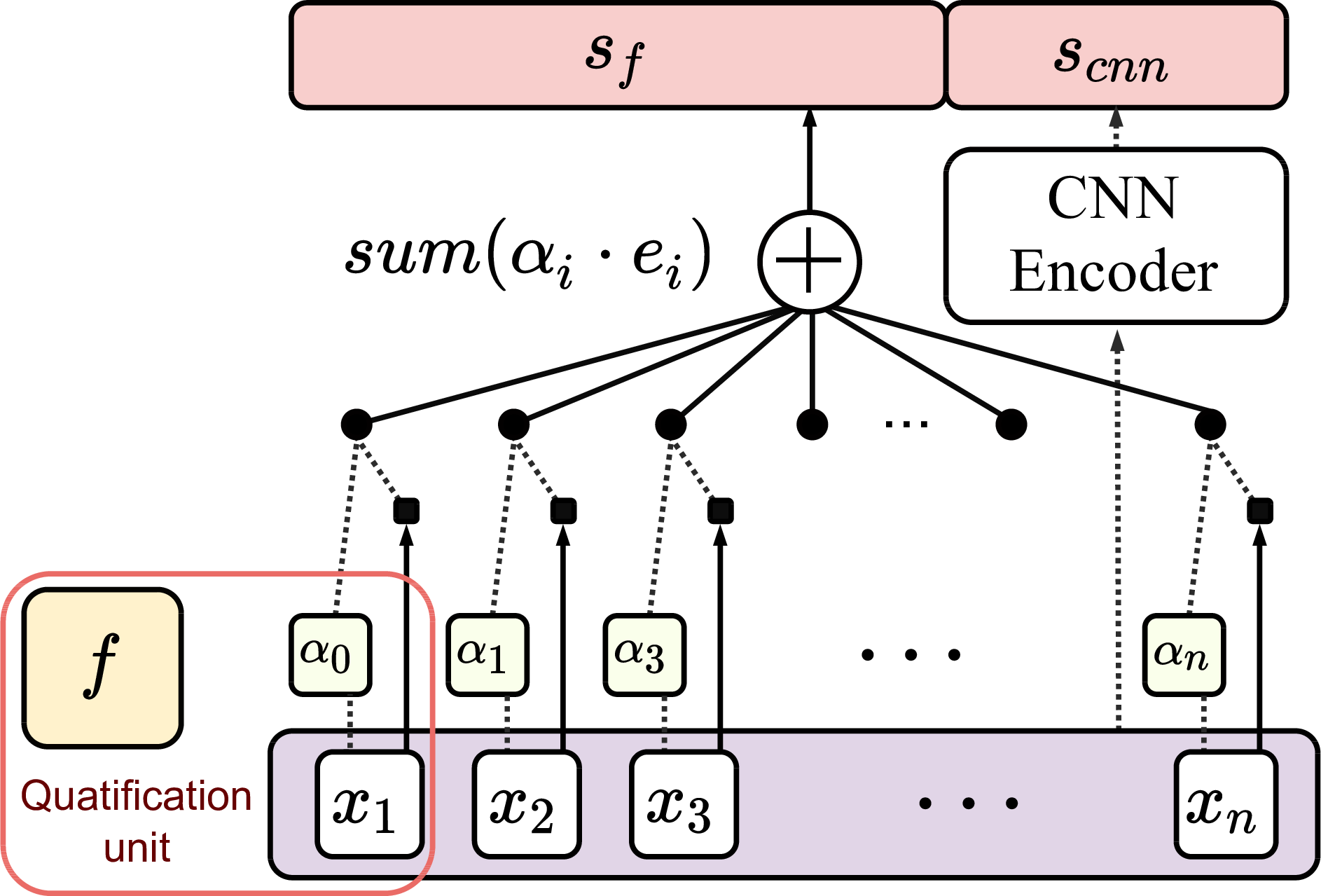}
        \caption{Context encoder architecture}
        \label{fig:mlp-attention-encoder}
    \end{subfigure} \hspace{0.2cm}
    \begin{subfigure}{.38\textwidth}
        \centering
        \includegraphics[width=\linewidth]{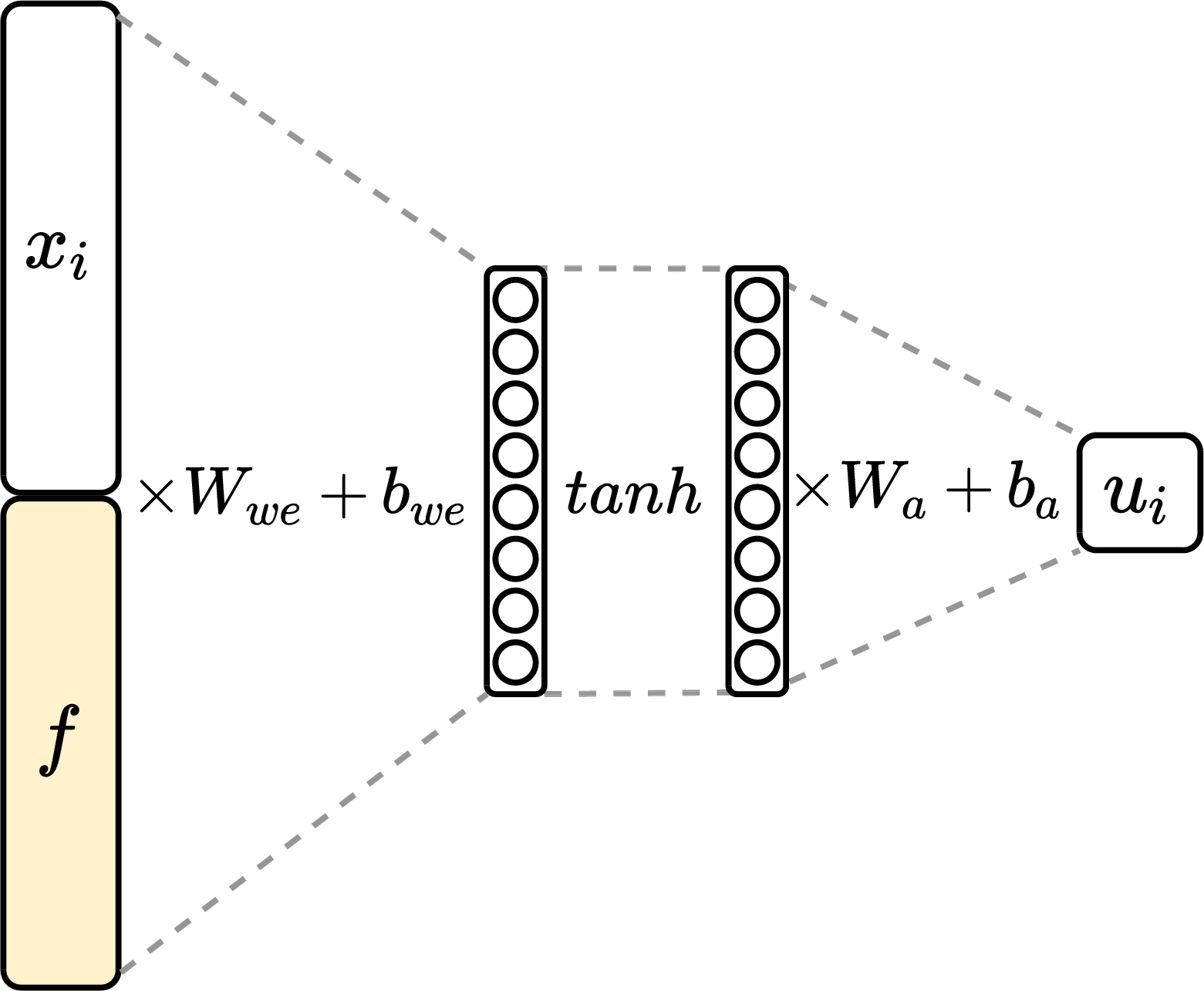}
        \caption{Quantification of term $\word{j}$ as its embedded representation $\embWord{j}$ relevance with respect to $\embFeature{}~\in~\featureSet{}$}
        \label{fig:mlp-weight-calc}
    \end{subfigure}
    \caption{\attCnn{} neural network~\cite{huang2016attention}}
    \label{fig:mlp-based-model}
\end{figure}

\begin{equation}
    \hidden{i} = \left[\embWord{i}, \embFeature{} \right]
    \label{eq:mlp_concat}
    \hspace{0.8cm}
    \hidden{i} \in \mathbb{R}^{2\cdot m}
\end{equation}

The quantification of the relevance of $\embWord{i}$ with  respect to $\embFeature{}$
is denoted as $u_{i}~\in~\mathbb{R}$ and calculated as follows (see Figure~\ref{fig:mlp-weight-calc}):
\begin{equation}
    u_{i} = W_a\left[\tanh(W_{we} \cdot h_{i} + b_{we})\right] + b_a \hspace{0.5cm}
    W_{we} \in \mathbb{R}^{2 \cdot m \times \mlpHidden}, \hspace{0.2cm}
    W_a \in \mathbb{R}^{\mlpHidden{}}
    \label{eq:mlp_weight}
\end{equation}

In Formula~\ref{eq:mlp_weight}, $W_{we}$ and $W_a$ 
correspond to the weight and attention matrices respectively, and 
$\mlpHidden{}$ corresponds to the size of the hidden representation in the weight matrix.
To deal with normalized weights within a context, we transform quantified values $u_{i}$ 
into probabilities $\alpha_{i}$ using $softmax$ operation (Formula~\ref{eq:softmax}).
We utilize Formula~\ref{eq:mlp_emb} to obtain attention-based context embedding $\featEmb{}$
of a context with respect to feature $\embFeature{}$:
\begin{equation}
\featEmb{} = \sum_{i=1}^{n} \embWord{i} \cdot \alpha_{i}
\hspace{1cm}
\featEmb{} \in \mathbb{R}^{\embSize{}}
\label{eq:mlp_emb}
\end{equation}

Applying Formula~\ref{eq:mlp_emb} towards each feature $\embFeature{j} \in \featureSet{}, \hspace{0.1cm} j \in \overline{1..\featuresCount{}}$
results in vector $\{\featEmb{}_j\}_{j=1}^{\featuresCount{}}$.
We use \textit{average-pooling} to transform the latter sequence into
single averaged vector
$s_{f} = \featEmb{}_{j} / [\sum_{j=1}^{k} \featEmb{}_{j} ]$.

We also utilize a CNN-based encoder (Figure~\ref{fig:mlp-attention-encoder})
to compete the context representation $s_{cnn} \in \mathbb{R}^{\convCount{}}$
, where \convCount{} is related to convolutional filters count~\cite{rusnachenko2018neural}.
The resulting context embedding vector 
$\sentenceEmbedding{}$
(size of
$\sentEmbSize{} = \featuresCount{} + \convCount{}$)
is a concatenation of 
$s_{f}$
and 
$s_{cnn}$.
\subsubsection*{IAN.}

\newcommand{\ianHidden}{\textbf{h}}
\newcommand{\ianC}{c}   
\newcommand{\ianF}{f}   

\newcommand{\Wc}{W_\ianC{}}
\newcommand{\Wf}{W_\ianF{}}

\newcommand{\Pc}{p_\ianC{}}
\newcommand{\Pf}{p_\ianF{}}

\newcommand{\Bc}{b_\ianC{}}
\newcommand{\Bf}{b_\ianF{}}

\newcommand{\Hc}[1]{h^{\ianC{}}_{#1}}
\newcommand{\Hf}[1]{h^{\ianF{}}_{#1}}

\newcommand{\Uc}[1]{u^{\ianC{}}_{#1}}
\newcommand{\Uf}[1]{u^{\ianF{}}_{#1}}

\newcommand{\AlphaC}[1]{\alpha_{#1}^\ianC{}}
\newcommand{\AlphaF}[1]{\alpha_{#1}^\ianF{}}

\newcommand{\Sc}{s_\ianC{}}
\newcommand{\Sf}{s_\ianF{}}

As a context encoder, 
a Recurrent Neural Network (\textsc{RNN}) model allows treating the context
$[\word{1}, \ldots, \word{\ctxSize{}}]$ as a sequence of terms
to generate a hidden representation, enriched with features of previously appeared terms.
In comparison with \cnn{}, 
the application of \textsc{rnn} allows keeping a history 
of the whole sequence while \cnn{}-based encoders remain limited by the window size.
The application of RNN towards a context and certain features appeared in it -- is another way
how the correlation of these both factors could be quantitatively measured~\cite{ma2017interactive}.

\begin{figure}[t]
    \begin{subfigure}{.59\textwidth}
        \centering
        \includegraphics[width=0.91\textwidth]{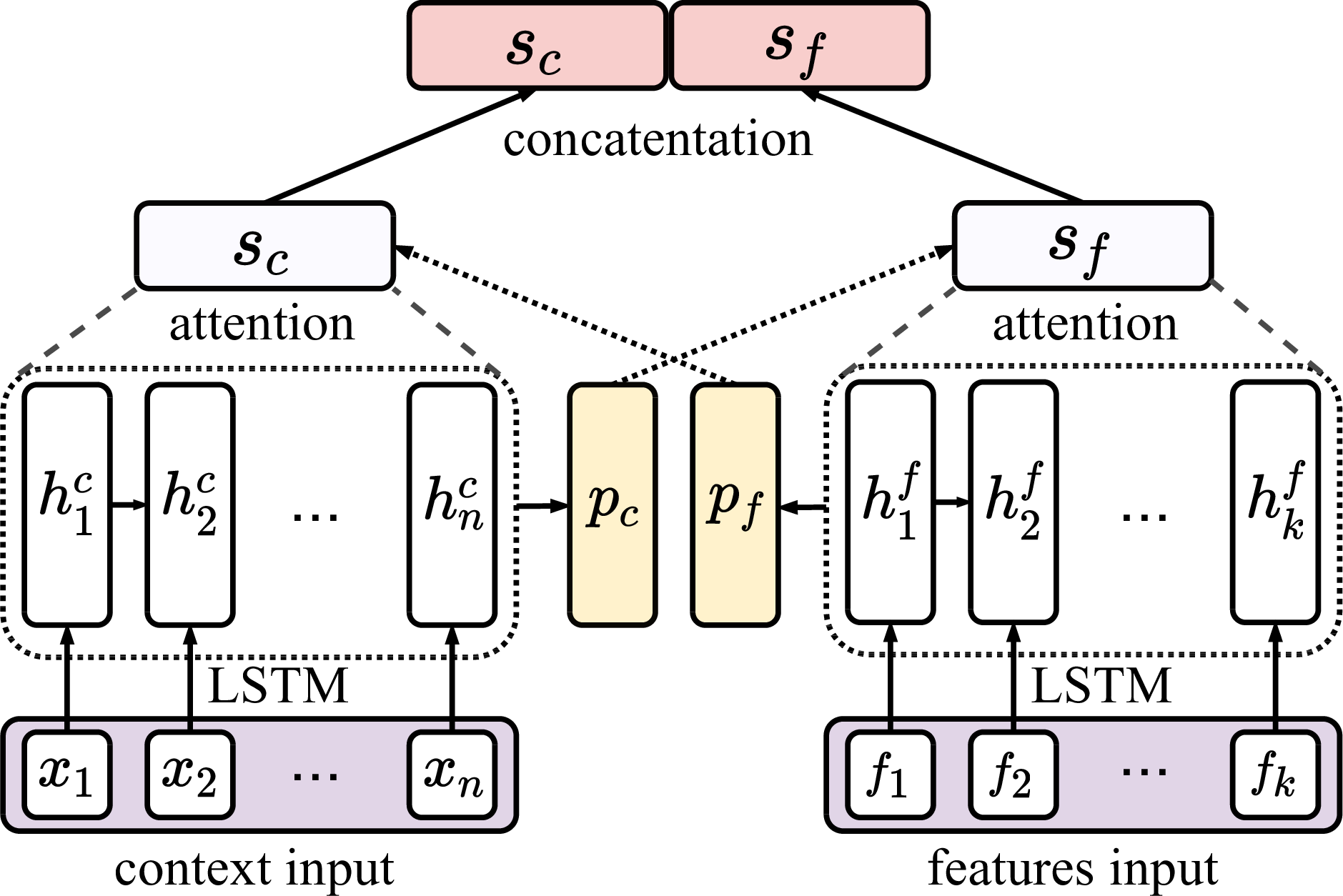}
        \caption{Context encoder architecture}
        \label{fig:ian-attention-encoder}
    \end{subfigure} \hspace{0.2cm}
    \begin{subfigure}{.34\textwidth}
        \centering
        \includegraphics[width=\textwidth]{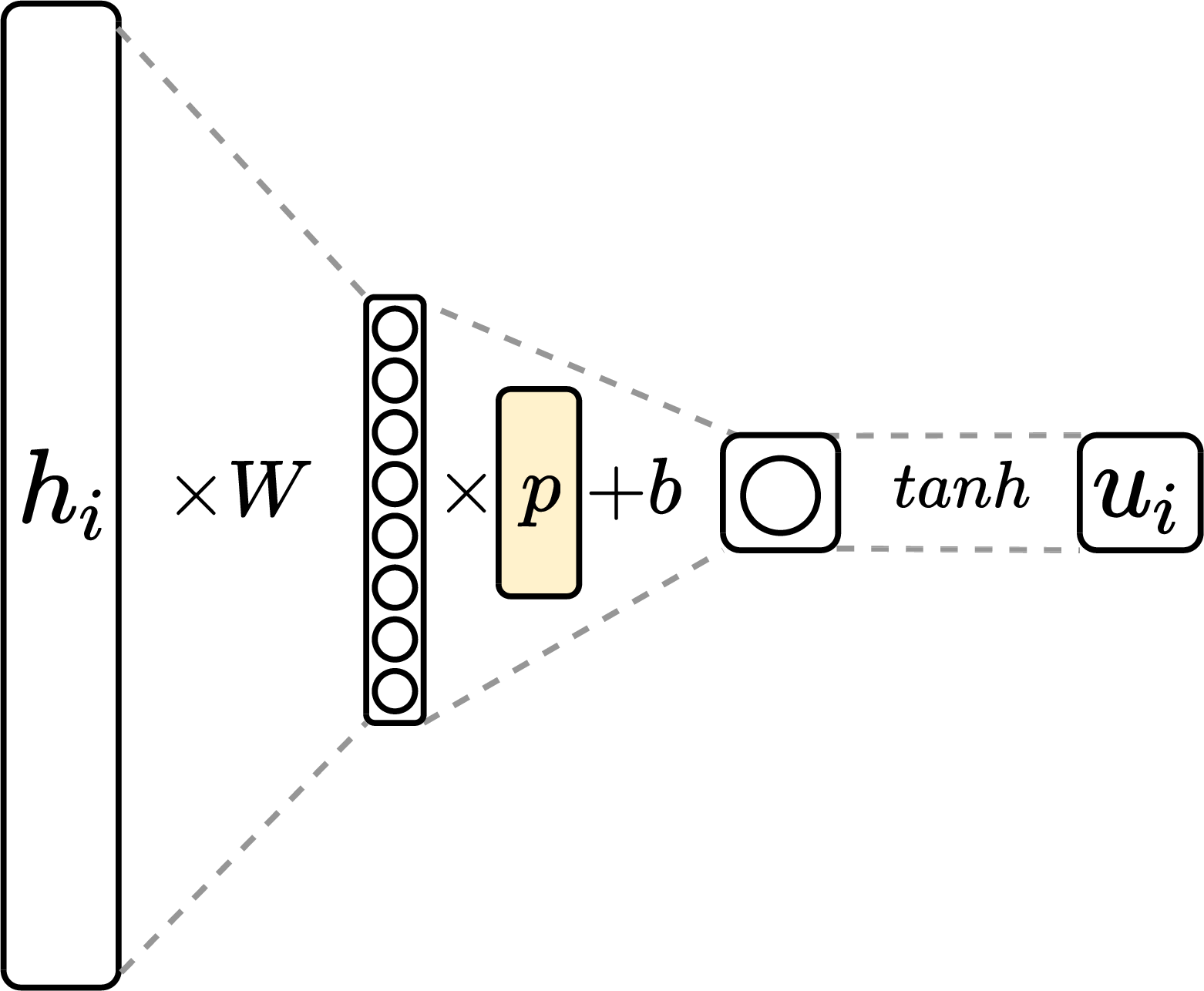}
        \caption{Quantification of term $\word{i}$ based on related hidden state $\hidden{i}$ with  respect to pooled
        representation $p \in \mathbb{R}^{\ianHidden}$}
        \label{fig:ian-weight-calc}
    \end{subfigure}
    \caption{Interactive Attention Network (\ian{})~\cite{ma2017interactive}}
    \label{fig:ian}
\end{figure}

Figure~\ref{fig:ian-attention-encoder} illustrates the  \ian{} architecture attention encoder.
The input assumes  separated sequences of embedded terms $\embSet{}$ and
embedded features $\featureSet{}$.
To learn the hidden term semantics for each input,
we utilize the \lstm{}~\cite{hochreiter1997long} recurrent neural network architecture, which addresses
learning long-term dependencies by avoiding gradient vanishing and expansion problems.
The calculation $\hidden{t}$ of $t$'th embedded term $\embWord{t}$ based
on prior state $\hidden{t-1}$, where the latter acts as a parameter of
auxiliary functions~\cite{hochreiter1997long}.
The application of \lstm{} towards the input sequences results in
$[\Hc{1},~\ldots,~\Hc{\ctxSize{}}]$ and
$[\Hf{1},~\ldots,~\Hf{\featuresCount{}}]$, where
$\Hc{i},~\Hf{j}~\in~\mathbb{R}^{\ianHidden{}}$
($i \in \overline{1..\ctxSize{}}, \hspace{0.2cm} j \in \overline{1..\featuresCount{}}$)
and $\ianHidden{}$ is the size of the hidden representation.
The quantification of input sequences is carried out in the following directions:
(\romannum{1}) feature representation with respect to context, and
(\romannum{2}) context representation with respect to features.
To obtain the representation of a hidden sequence, we utilize \textit{average-pooling}.
In Figure~\ref{fig:ian-attention-encoder},
$\Pf{}$ and $\Pc{}$ denote a hidden representation of features and context respectively.
Figure~\ref{fig:ian-weight-calc} illustrates the quantification computation of a hidden state $\hidden{t}$ with
respect to $p$:
\begin{equation}
\begin{split}
    \Uc{i} = \tanh(\Hc{i} \cdot \Wf{} \cdot \Pf{} + \Bf{}) \hspace{1cm}
    \Wf{} \in \mathbb{R}^{\ianHidden \times \ianHidden}, \hspace{0.2cm}
    \Bf{} \in \mathbb{R}, \hspace{0.2cm}
    i \in \overline{1 .. \ctxSize{}} \\
    \Uf{j} = \tanh(\Hf{j} \cdot \Wc{} \cdot \Pc{} + \Bc{}) \hspace{1cm}
    \Wc{} \in \mathbb{R}^{\ianHidden \times \ianHidden}, \hspace{0.2cm}
    \Bc{} \in \mathbb{R}, \hspace{0.2cm}
    j \in \overline{1 .. \featuresCount{}}
    \label{eq:ian-weight}
\end{split}
\end{equation}

In order to deal with normalized weight vectors
$\AlphaF{i}$
and 
$\AlphaC{j}$,
we utilize the $softmax$ operation for $\Uf{}$ and $\Uc{}$ respectively (Formula~\ref{eq:softmax}).
The resulting context vector $\sentenceEmbedding{}$
(size of $\sentEmbSize{}=2\cdot\ianHidden{}$)
is a concatenation of weighted context $\Sc{}$ and features $\Sf{}$ representations:
\begin{equation}
    \Sc{} = \sum_{i=1}^{\ctxSize{}} \AlphaC{i} \cdot \Hc{i} \hspace{1cm}
    \Sf{} = \sum_{j=1}^{\featuresCount{}} \AlphaF{j} \cdot \Hf{j}
\end{equation}
\subsection{Self Attentive Context Encoders}
\label{sec:self-based}

In section~\ref{sec:feature-based} the application of attention in context embedding fully relies on the sequence of predefined features.
The quantification of context terms is performed towards each feature.
In turn, the \textit{self-attentive} approach assumes to quantify a context with respect to an
abstract parameter.
Unlike quantification methods in feature-attentive embedding models,
here the latter is replaced with a hidden state (parameter $w$, see Figure~\ref{fig:self-weight-calc}),
which modified during the training process.

\newcommand{\selfHidden}{\textbf{h}}

\newcommand{\selfHiddenVector}{w}

\newcommand{\selfAttHidden}{H_a}

\begin{figure}[!tp]
    \captionsetup[subfigure]{position=b}
    \begin{subfigure}{.585\textwidth}
        \centering
        \includegraphics[width=0.85\textwidth]{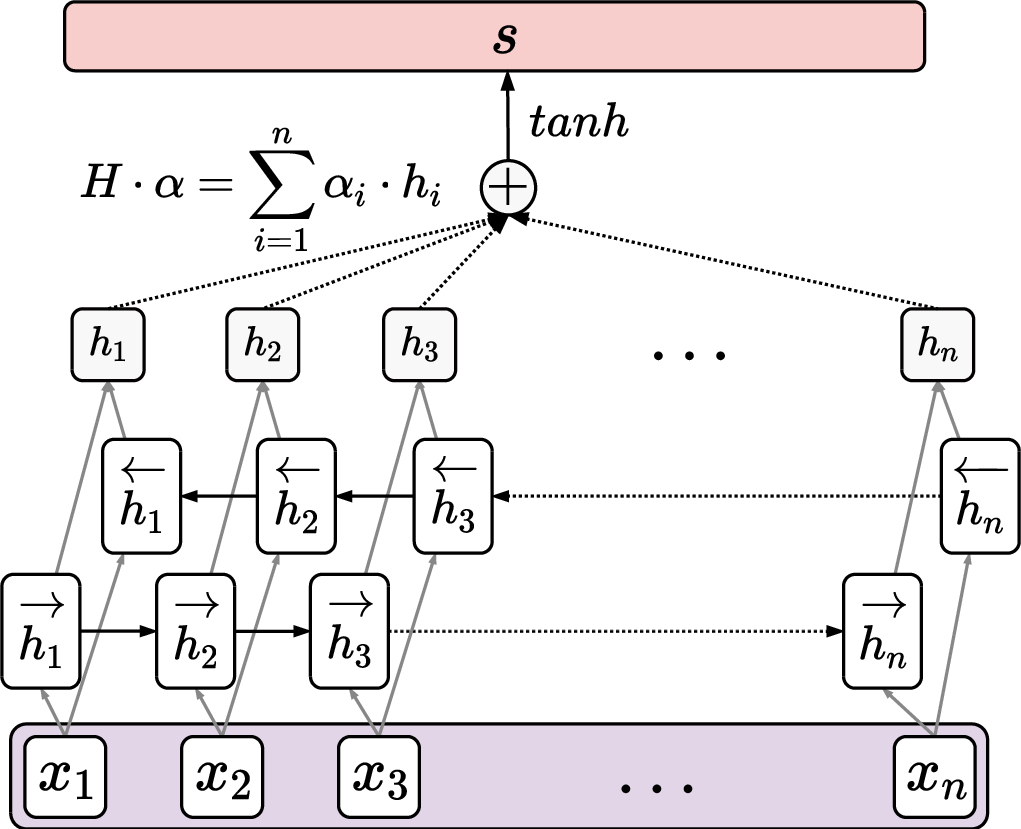}
        \caption{Context encoder architecture}
        \label{fig:self-attention-encoder}
    \end{subfigure} \hspace{0.2cm}
    \begin{subfigure}{.35\textwidth}
        \centering
        \includegraphics[width=\linewidth]{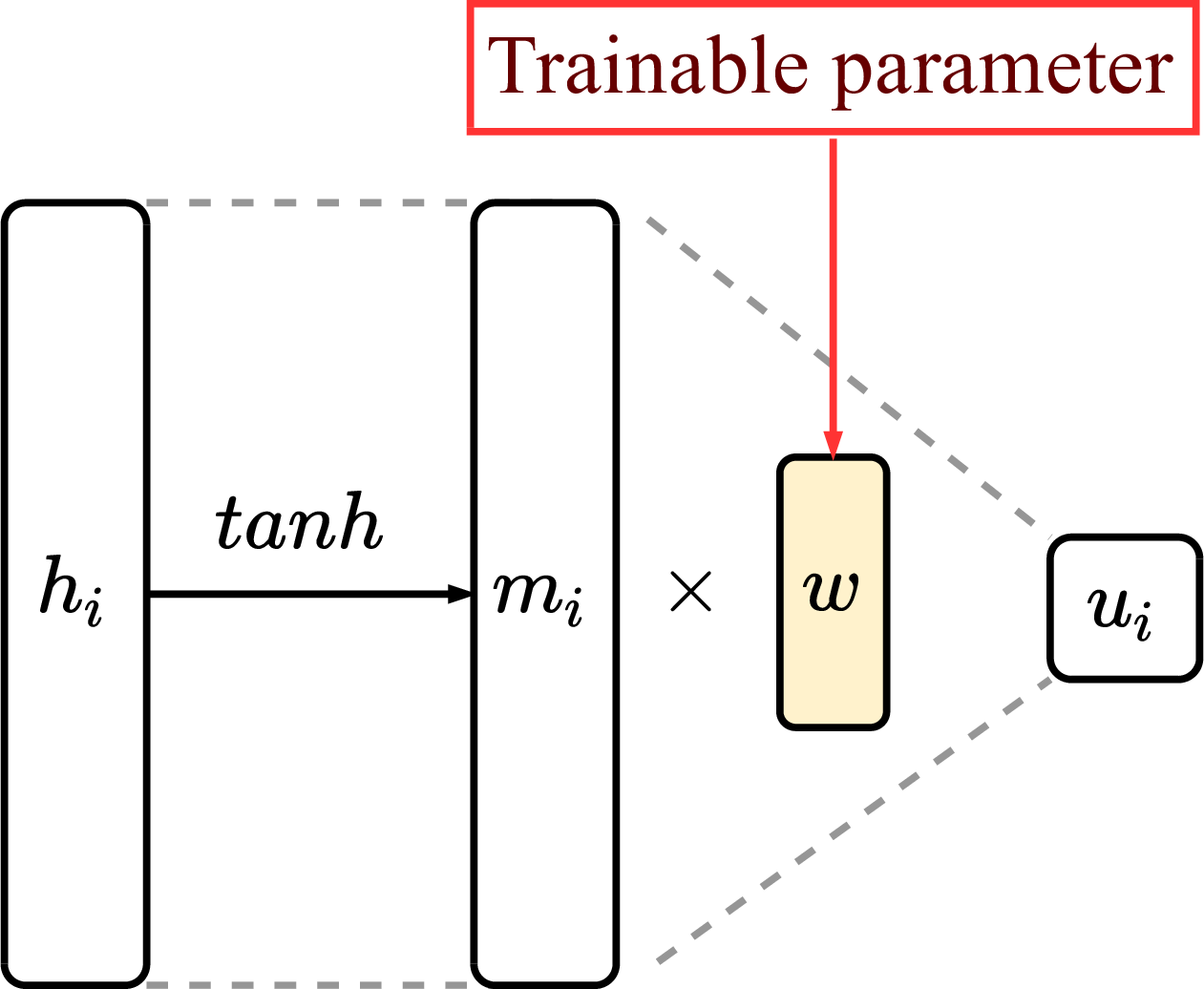}
        \caption{Quantification of $\hidden{j}$ with  respect to parameter $\selfHiddenVector{}$~\cite{zhou2016attention};
        $\selfHiddenVector{}$ represents a hidden vector which modifies during model training process}
        \label{fig:self-weight-calc}
    \end{subfigure}
    \caption{Attention-based bi-directional \lstm{} neural network (\bilstmPZhou{})~\cite{zhou2016attention}}
    \label{fig:self-based-model}
\end{figure}

Figure~\ref{fig:self-attention-encoder} illustrates the bi-directional RNN-based
self-attentive context encoder architecture.
We utilize bi-directional \lstm{} (\bilstm{}) to obtain a pair of sequences
$\overrightarrow{\hidden{}}$
and
$\overleftarrow{\hidden{}}$ \hspace{0.5mm}
($\overrightarrow{\hidden{i}}, \overleftarrow{\hidden{i}} \in \mathbb{R}^{\selfHidden{}}$).
The resulting context representation $H=[\hidden{1},~\ldots,~\hidden{\ctxSize{}}]$ is
composed as the concatenation of bi-directional sequences elementwise:
$\hidden{i} = \overrightarrow{\hidden{i}} + \overleftarrow{\hidden{i}}, \hspace{0.1cm}
i \in \overline{1..\ctxSize{}}$.
The quantification of hidden term representation $\hidden{i} \in \mathbb{R}^{2 \cdot \selfHidden{}}$
with respect to $w \in \mathbb{R}^{2 \cdot \selfHidden{}}$ is
described in formulas~\ref{eq:self-q1}-\ref{eq:self-q2} and illustrated in Figure~\ref{fig:self-weight-calc}.
\begin{equation}
    m_i = \tanh(\hidden{i})
    \label{eq:self-q1}
\end{equation}
\begin{equation}
    u_i = m_i^T \cdot w
    \label{eq:self-q2}
\end{equation}

We apply the $softmax$ operation towards $u_i$ to obtain vector of normalized weights
$\alpha~\in~\mathbb{R}^n$.
The resulting context embedding vector
$\sentenceEmbedding{}$
(size of $\sentEmbSize{} = 2 \cdot \selfHidden{}$)
is an activated weighted sum of each parameter of  context hidden states:
\begin{equation}
    \sentenceEmbedding{} = tanh(H \cdot \alpha)
\end{equation}


\section{Model Details}
\label{sec:model_details}

\subsubsection*{Input Embedding Details}
\label{sec:input-embedding-details}

\newcommand{\featureSize}{5}
\newcommand{\embeddingWindowSize}{20}
\newcommand{\embeddingVectorSize}{1000}
\newcommand{\wordEmbedding}{$M_{word}$}

\newcommand{\featurePOS}{$v_\textsc{pos}$}
\newcommand{\featureDistObj}{$v_{\textsc{d}\mhyphen obj}$}
\newcommand{\featureDistSubj}{$v_{\textsc{d}\mhyphen subj}$}
\newcommand{\featureSynDistObj}{$v_{\textsc{sd}\mhyphen obj}$}
\newcommand{\featureSynDistSubj}{$v_{\textsc{sd}\mhyphen subj}$}
\newcommand{\featurePolarity}{$v_{A0\to A1}$}

We provide embedding details of context term groups described in Section~\ref{sec:model}.
For \wordsGroup{} and \framesGroup{}, we look up for vectors in precomputed and publicly available model\footnote{
\url{http://rusvectores.org/static/models/rusvectores2/news_mystem_skipgram_1000_20_2015.bin.gz}}
\wordEmbedding{} based on news articles with 
window size of $\embeddingWindowSize{}$,
and vector size of $\embeddingVectorSize{}$.
Each term that is not presented in the model we treat as a sequence of \textit{parts} ($n$-grams)
and look up for related vectors in~\wordEmbedding{} to complete an averaged vector.
For a particular part, we start with a trigram ($n=3$) and decrease $n$ until the related $n$-gram is found.
For masked entities (\maskE{}, \maskEObj{}, \maskESubj{}) and \tokenGroup{}, each element embedded with a randomly initialized vector with size of 1000.

Each context term has been additionally expanded with the following parameters:
\begin{itemize}
    \item Distance embedding~\cite{rusnachenko2018neural} (\featureDistObj{}, \featureDistSubj{})
    -- is vectorized distance in terms from attitude participants of entry pair (\maskEObj{} and \maskESubj{} respectively) 
    to a given term;
    \item Closest to synonym distance embedding (\featureSynDistObj{}, \featureSynDistSubj{})
    is a vectorized absolute distance in terms from
    a given term towards the nearest entity, synonymous to \maskEObj{} and \maskESubj{} respectively; 
    \item Part-of-speech embedding (\featurePOS{})
    is a vectorized tag  for \wordsGroup{} (for terms of other groups considering <<unknown>> tag);
    \item \polarity{} polarity embedding (\featurePolarity{}) is
    a vectorized <<positive>> or <<negative>> value for frame entries whose description in \rusentiframes{} provides the corresponding polarity
    (otherwise considering <<neutral>> value);
    polarity is inverted when an entry has \begin{otherlanguage*}{russian}"не"\end{otherlanguage*} (not) preposition.
\end{itemize}


\subsubsection*{Training}

\newcommand{\lossF}[1]{L_{#1}}

This process assumes hidden parameter optimization of a given model.
We utilize an algorithm described in~\cite{rusnachenko2018neural}.
The input is organized in minibatches, where minibatch yields of $\bagsCount{}$ \textit{bags}.
Each bag has a set of $\sentencesCount{}$ pairs
$\left<\embSet{}_j, y_j\right>_{j=1}^{\sentencesCount{}}$,
where each pair is described by an input embedding $\embSet{}_j$
with the related label $y_j\in\mathbb{R}^\classesCount{}$.
The training process is iterative, and each iteration includes the following steps:
\begin{enumerate}
    \item Composing a minibatch of $\bagsCount{}$ bags of size $\sentencesCount{}$;
    \item Performing forward propagation through the network which results in
    a vector (size of $q = \bagsCount{} \cdot \sentencesCount{}$) 
    of outputs $o_k\in\mathbb{R}^\labelsCount{}$;
    \item Computing cross entropy loss for output: 
            $\lossF{k} = \sum\limits_{j=1}^\labelsCount{} \log p(y_i|o_{k,j}; \theta), \hspace{0.1cm} k \in \overline{1..q}$;
    \item Composing cost vector $\{cost_i\}_{i=1}^{\bagsCount{}}$, $cost_i = \max\left[\lossF{(i-1) \cdot \sentencesCount} \hspace{0.1cm} .. \hspace{0.1cm} \lossF{i\cdot \sentencesCount}\right)$ to update hidden variables set; 
        $cost_i$  is a maximal loss within \textit{i}'th bag;
\end{enumerate}

\subsubsection*{Parameters settings}

\newcommand{\bagsCountValue}{$2$}
\newcommand{\ctxPerBag}{$3$}
\newcommand{\termsPerContext}{50}
\newcommand{\framesPerContext}{$5$}
\newcommand{\lstmHidden}{$128$}

\newcommand{\cnnWindowSizeValue}{$3$}
\newcommand{\cnnFiltersCountValue}{$300$}

\newcommand{\dropoutKeepProb}{$0.8$}

The minibatch size ($\bagsCount{}$) is set to $\bagsCountValue{}$, 
where contexts count per bag $\sentencesCount{}$ is set to $\ctxPerBag{}$.
All the sentences were limited by $\termsPerContext{}$ terms.
For embedding parameters
(\featureDistObj{},
\featureDistSubj{},
\featureSynDistObj{},
\featureSynDistSubj{},
\featurePOS{},
\featurePolarity{}),
we use randomly initialized vectors with size of~$\featureSize{}$.
For \cnn{} and \pcnn{} context encoders, 
the size of convolutional window
and filters count (\convCount{})
were set to \cnnWindowSizeValue{} and \cnnFiltersCountValue{} respectively. 
As for parameters related to sizes of hidden states in Section~\ref{sec:model}:
$\mlpHidden{}=10$, 
$\ianHidden{}=\lstmHidden{}$. 
For feature attentive encoders, we keep frames in order of their appearance in context and limit $\featuresCount{}$~by~\framesPerContext{}.
We utilize the AdaDelta optimizer with parameters $\rho=0.95$ and $\epsilon=10^{-6}$~\cite{zeiler2012adadelta}.
To prevent models from overfitting, 
we apply $dropout$ towards the output
with keep probability set to~\dropoutKeepProb{}.
We use Xavier weight initialization to setup initial values for hidden states~\cite{glorot2010understanding}.

\begin{table}[!tp]
\centering
\begin{tabular}{l|c|ccc||c}
\hline
        & & & & & \\ [-0.95em]
                           Model & ${F1}_{avg}$ & ${F1}_{cv}^1$ & ${F1}_{cv}^2$  & ${F1}_{cv}^3$  & ${F1}$\testSc{}  \\
        \hline
        \bilstmPZhou{} &   \bf{0.314} & \textbf{0.35} &            0.27 &      \textbf{0.32}      &   \textbf{0.35} \\
        \bilstmZYang{} &        0.292 &          0.33 &            0.25 &             0.30        &        0.33 \\
        \bilstm{}      &        0.286 &          0.32 &            0.26 &             0.28        &        0.34 \\
        \hline
        \ianEf{}        &       0.289 &          0.31 &            0.28 &             0.27        &        0.32 \\
        \ianEnds{}      &       0.286 &          0.31 &            0.26 &             0.29        &        0.32 \\
        \lstm{}         &       0.284 &          0.28 &            0.27 &             0.29        &        0.32 \\
        \hline
        \pcnnEnds{}     &       0.297 &          0.32 &    \textbf{0.29} &           0.28        & \textbf{0.35} \\
        \pcnnEf{}       &       0.289 &          0.31 &            0.25  &           0.31        &         0.31 \\
        \pcnn{}         &       0.285 &          0.29 &            0.27  &           0.30        &         0.32 \\
    \hline
    \end{tabular}
    \caption{Three class context classification results by $F1$ 
             measure (\rusentrel{}~dataset);
             Columns from left to right:
             (\romannum{1}) average value in CV-3 experiment ($F1_{avg}$) with results on each split
             ($F1_{cv}^i, \hspace{0.1cm} i \in \overline{1..3}$);
             (\romannum{2}) results on \train{}/\test{} separation ($F1_\test{}$)}
    \label{tab:results}
\end{table}

\section{Experiments}
\label{sec:experiments}

We conduct experiments with the ~\rusentrel{}\footnote{\rusentrelLink{}} corpus in following formats:
\begin{enumerate}
    \item Using 3-fold cross-validation (CV),
    where all folds are equal in terms of the number of sentences;
    \item Using predefined \train{}/\test{} separation\footnote{\url{https://miem.hse.ru/clschool/results}}.
\end{enumerate}

In order to evaluate and assess attention-based models,
we provide a list of baseline models.
These are independent encoders described in Sections~\ref{sec:feature-based} and~\ref{sec:self-based}:
\pcnn{}~\cite{rusnachenko2018neural},
\lstm{},
\bilstm{}.
In case of models with feature-based attentive encoders (\ian{}$_{*}$, \pcnn{}$_{*}$)
we experiment with following feature sets:
attitude participants only (\featureEnds{}), and
frames with attitude participants (\featureEf{}).
For self-based attentive encoders
we experiment with
\bilstmPZhou{}~(Section~\ref{sec:self-based})
and
\bilstmZYang{} -- is a bi-directional \lstm{} model with word-based attentive encoder of
\han{} model~\cite{yang2016hierarchical}.

Table~\ref{tab:results} provides related results. 
For evaluating models in this task,
we adopt macroaveraged F1-score (\fmeasure{})
over documents.
F1-score is considered averaging of the positive and negative class.
We measure \fmeasure{} on train part every \epochsToTest{} epochs.
The number of epochs was limited by \epochsCount{}.
The training process terminates when \fmeasure{} on train part become greater than $0.85$.
Analyzing $F1_\test{}$ results
it is quite difficult to demarcate attention-based models from baselines except \bilstmPZhou{} and \pcnnEnds{}.
In turn, average results by \fmeasure{} in the case of CV-3 experiments
illustrate the effectiveness of attention application.
The average increase in the performance of such models over related baselines is as follows:
$\pcnnIncrease{}$\% (\pcnn{}$_{*}$),
$\ianIncrease{}$\% (\ian{}$_{*}$),
and
$\bilstmIncrease{}$\% (\bilstmPZhou{}, \bilstmZYang{})
by
\fmeasure{}.
The greatest increase in \improvePZhou{}\% by \fmeasure{} is achieved by \bilstmPZhou{} model.

\newcommand{\modelA}{(1)}
\newcommand{\modelB}{(2)}
\newcommand{\modelC}{(3)}

\newcommand{\modelAName}{\pcnnEf{}}
\newcommand{\modelBName}{\ianEf{}}
\newcommand{\modelCName}{\bilstmPZhou{}}

\begin{figure}[!t]
    \centering
    \begin{tabular}{@{}cccc@{}}
        \includegraphics[width=.31\textwidth]{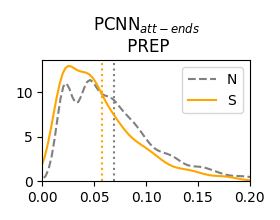} &
        \includegraphics[width=.31\textwidth]{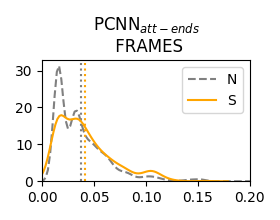} &
        \includegraphics[width=.31\textwidth]{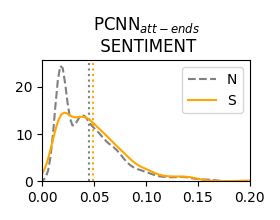} \\ [-1em]

        \includegraphics[width=.31\textwidth]{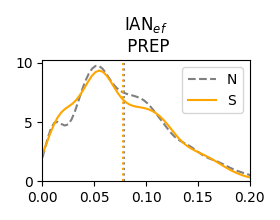} &
        \includegraphics[width=.31\textwidth]{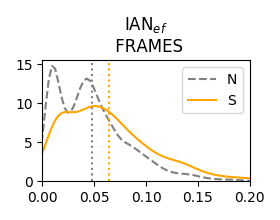} &
        \includegraphics[width=.31\textwidth]{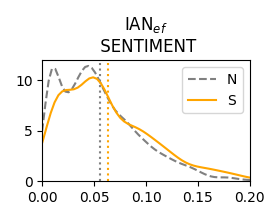} \\ [-1em]

        \includegraphics[width=.31\textwidth]{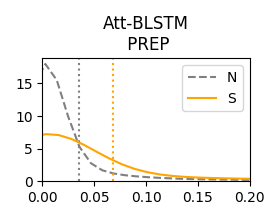} &
        \includegraphics[width=.31\textwidth]{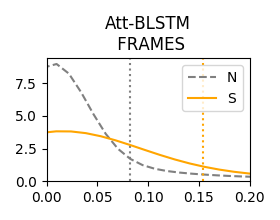} &
        \includegraphics[width=.31\textwidth]{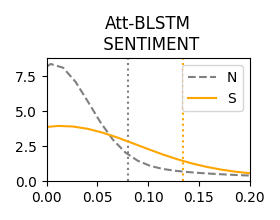}
    \end{tabular}
    \caption{Kernel density estimations (KDE) of context-level weight distributions of 
    term groups (from left to right: \prepGroup{}, \framesGroup{}, \sentGroup{})
    across
    \textit{neutral} (\texttt{N})
    and
    \textit{sentiment} (\texttt{S})
    context sets for models:
    \modelAName{}, \modelBName{}, \modelCName{};
    the probability range (x-axis) scaled to $[0, 0.2]$;
    vertical 
    lines indicate expected values of distributions}
    \label{fig:dist-analysis}
\end{figure}

\section{Analysis of Attention Weights}

\newcommand{\IANClass}{\textsc{IAN}}

According to Sections~\ref{sec:feature-based} and~\ref{sec:self-based},
attentive embedding models perform the quantification of terms in the context.
The latter results in the probability distribution of weights\footnote{
We consider and analyze only context weights in case of \IANClass{} models}
across the terms mentioned in a context.

We utilize the \test{}  part of the \rusentrel{} dataset (Section~\ref{sec:experiments}) for analysis of weight distribution of \framesGroup{} group,
declared in Section~\ref{sec:model}, across all input contexts.
We also introduce two extra groups utilized in the analysis by separating the subset of \wordsGroup{} into
prepositions (\prepGroup{}) and terms appeared in \rusentilex{} lexicon (\sentGroup{}) described in Section~\ref{sec:data}.

\newcommand{\sD}[1]{$\rho_S^{#1}$}
\newcommand{\nD}[1]{$\rho_N^{#1}$}

The \textit{context-level weight} of a group is a weighted sum of terms which both appear
in the context and belong the corresponding term group.
Figure~\ref{fig:dist-analysis} illustrates the weight distribution plots,
where the models are organized in rows,
and the columns correspond to the term groups.
Each plot combines distributions of context-levels weights across:
\begin{itemize}
\item \textbf{Neutral contexts} -- contexts, labeled as \textbf{neutral};
\item \textbf{Sentiment contexts}  -- contexts, labeled with \textbf{positive or negative} labels.
\end{itemize}

In Figure~\ref{fig:dist-analysis} and further, the distribution of context-level weights across neutral
(<<\texttt{N}>> in legends) and
sentiment contexts
(<<\texttt{S}>> in legends)
denoted as \nD{g} and \sD{g} respectively.
The rows in Figure~\ref{fig:dist-analysis} correspond to the following models:
\modelA{}~\modelAName{},
\modelB{}~\modelBName{},
\modelC{}~\modelCName{}.
Analyzing prepositions (column~1) it is possible to see the lack of differences in quantification
between the \nD{\prepGroup{}} and \sD{\prepGroup{}} contexts in the case of the
models \modelA{} and \modelB{}.
Another situation is in case of the model \modelC{}, where related terms in sentiment contexts are higher
quantified than in neutral ones.
\framesGroup{} and \sentGroup{} groups
are slightly higher quantified in sentiment contexts than in neutral one in the case of models
\modelA{} and \modelB{},
while \modelC{} illustrates a significant discrepancy.

\begin{figure}[t]
    \centering
    \includegraphics[width=\textwidth]{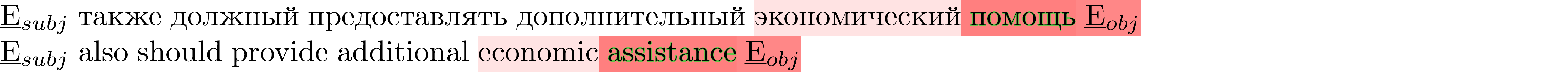} \\
    \vspace{0.1cm}
    \includegraphics[width=\textwidth]{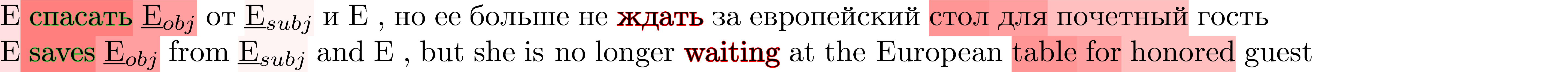} \\
    \vspace{0.1cm}
    \includegraphics[width=\textwidth]{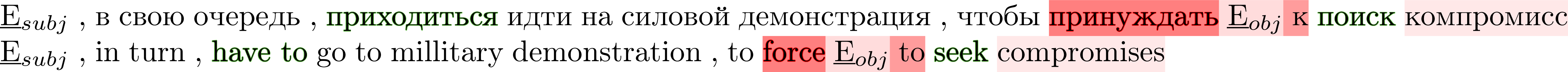} \\
    \caption{Weight distribution visualization for model \modelCName{} on sentiment contexts; 
    for visualization purposes,
    weight of each term is normalized by maximum in context}
    \label{fig:heatmap}
\end{figure}

Overall, model \modelCName{} stands out among others
both in terms of results (Section~\ref{sec:experiments})
and it illustrates the greatest discrepancy between \nD{} and \sD{}
across all the groups presented in the analysis (Figure~\ref{fig:dist-analysis}).
We assume that the latter is  achieved due to the following factors:
(\romannum{1}) application of bi-directional \lstm{} encoder;
(\romannum{2})  utilization of a single trainable vector ($w$) in the quantification process
(Figure~\ref{fig:self-weight-calc})
while the models of other approaches
(\attCnn{}, 
\ian{}, and 
\bilstmZYang{})
depend on fully-connected layers.
Figure~\ref{fig:heatmap} shows examples of those sentiment contexts
in which the weight distribution is the largest  among the \framesGroup{} group.
These examples are the case when both frame and attention masks convey context meaning.

\section*{Conclusion}

In this paper, we study the attention-based models, aimed to extract sentiment attitudes from analytical articles. 
The described models should classify a context with an attitude mentioned in it onto the following classes:
positive, negative, neutral.
We investigated two types of attention embedding approaches:
(\romannum{1}) feature-based,
(\romannum{2}) self-based.
We conducted experiments on  Russian analytical texts of the \rusentrel{} corpus and provide the  analysis of the results.
According to the latter, 
the advantage of attention-based encoders
over non-attentive was shown by the variety in weight distribution of 
certain term groups between sentiment and non-sentiment contexts.
The application of attentive context encoders illustrates the classification improvement in
\improveRange{}\% range by \fmeasure{}.
\newpage

\bibliographystyle{splncs04}
\bibliography{nldb2020}

\end{document}